\pdfoutput=1

\documentclass[11pt]{article}
\usepackage{natbib}
\usepackage{enumitem}
\usepackage{makecell}
\usepackage{lipsum}
\usepackage{nccmath} 
\usepackage{longtable}

\usepackage{booktabs} 
\usepackage{array} 

\usepackage[final]{acl}

\usepackage{times}
\usepackage{latexsym}

\usepackage[T1]{fontenc}

\usepackage[utf8]{inputenc}

\usepackage{microtype}

\usepackage{inconsolata}
\usepackage{subfig}

\usepackage{graphicx}
\usepackage[most]{tcolorbox}
\definecolor{imagetext}{RGB}{85, 59, 45} 

\usepackage{amsmath} 
\usepackage{amssymb}
\usepackage{booktabs}
\usepackage{multirow}
\usepackage{colortbl}
\usepackage{array}
\usepackage{pifont}
\usepackage{graphicx}

\usepackage{svg}
\usepackage{url}
\usepackage{color}
\usepackage{array}
\usepackage{adjustbox}

\definecolor{promptbackground}{RGB}{249, 248, 245} 
\definecolor{prompttext}{RGB}{95, 84, 73}          

\title{Rethinking Text-based Protein Understanding: Retrieval or LLM?}

\author{
 \textbf{Juntong Wu\textsuperscript{1, 2 \footnotemark[1]} },
 \textbf{Zijing Liu\textsuperscript{2 \footnotemark[1]} },
 \textbf{He Cao\textsuperscript{2 \footnotemark[1]} \footnotemark[2]},
 \textbf{Hao Li\textsuperscript{1, 2}},
 \textbf{Bin Feng\textsuperscript{2}},
\\
 \textbf{Zishan Shu\textsuperscript{1}},
 \textbf{Ke Yu\textsuperscript{1 \footnotemark[3]}},
 \textbf{Li Yuan \textsuperscript{1 \footnotemark[3]}},
 \textbf{Yu Li\textsuperscript{2 \footnotemark[3]}},
\\
\\
 \textsuperscript{1} Peking University, Shenzhen Graduate School \\
 \textsuperscript{2} International Digital Economy Academy (IDEA)
\\
 \small{
   \textbf{Correspondence:} \href{mailto:yuke.sz@pku.edu.cn}{yuke.sz@pku.edu.cn},
   \href{mailto:yuanli-ece@pku.edu.cn}{yuanli-ece@pku.edu.cn},
   \href{mailto:liyu@idea.edu.cn}{liyu@idea.edu.cn}
 }
}

\begin{document}
\maketitle

\footnotetext[1]{Equal contribution.} \footnotetext[2]{Project Lead.} 
\footnotetext[3]{Corresponding author.}

\begin{abstract}

In recent years, protein-text models have gained significant attention for their potential in protein generation and understanding. Current approaches focus on integrating protein-related knowledge into large language models through continued pretraining and multi-modal alignment, enabling simultaneous comprehension of textual descriptions and protein sequences. 
Through a thorough analysis of existing model architectures and text-based protein understanding benchmarks, we identify significant data leakage issues present in current benchmarks. Moreover, conventional metrics derived from natural language processing fail to assess the model's performance in this domain accurately. To address these limitations, we reorganize existing datasets and introduce a novel evaluation framework based on biological entities. 
Motivated by our observation, we propose a retrieval-enhanced method, which significantly outperforms fine-tuned LLMs for protein-to-text generation and shows accuracy and efficiency in training-free scenarios. Our code and data can be seen in \href{https://github.com/IDEA-XL/RAPM}{https://github.com/IDEA-XL/RAPM}.

\end{abstract}




\begin{figure}[!ht]
    \centering
\includegraphics[width=\linewidth]{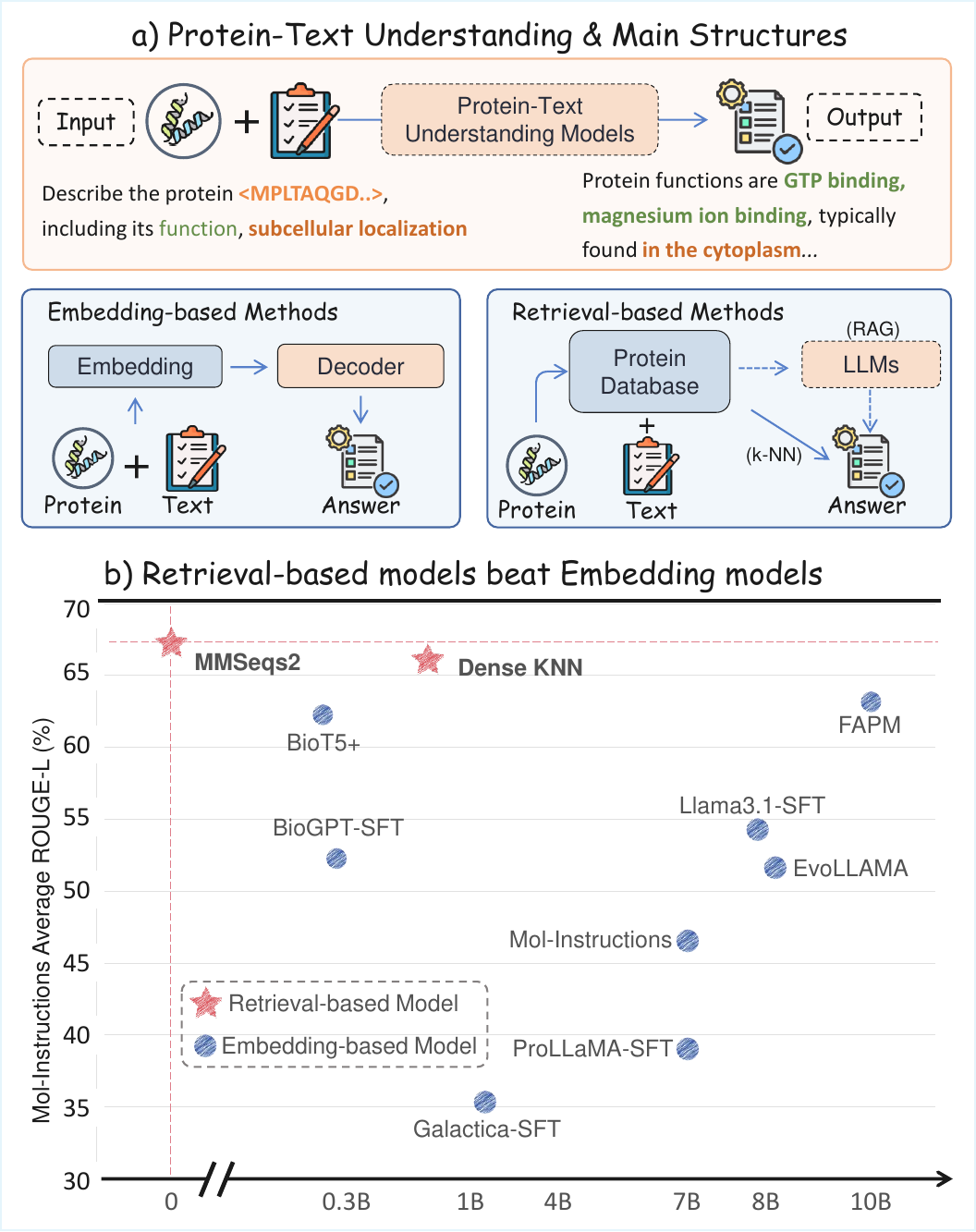}
    \vspace{-0.5cm}
    \caption{a) Protein understanding tasks, and LLM-based and retrieval-based methods for this task. b) The performance of existing methods in protein understanding tasks. Retrieval methods based on protein embeddings or sequences outperform LLM-based approaches.}
    \vspace{-0.4cm}
    \label{fig: teaser}
\end{figure}


\section{Introduction}







In recent years, large language models (LLMs) have achieved remarkable success across diverse domains~\citep{brown2020language, touvron2023llama, wei2022emergent, sallam2023utility,li2025decoupled}.
To further enhance the ability of LLMs in understanding domain-specific data (e.g., chemistry, biology), researchers have extended LLMs into the multi-modal domain, giving rise to multi-modal large language models (MLLMs)~\citep{liu2023visual, cao2023instructmol, maaz2023video}. 
Unlike traditional LLMs, which process single textual modality, MLLMs integrate multiple modalities, such as images, text, and graphs, by aligning them within a unified framework. This is typically accomplished using unimodal encoders for each input type and cross-modal projectors that map different modalities into a shared embedding space~\citep{zhu2023minigpt}. As a result, MLLMs enable sophisticated cross-modal reasoning, paving the way for applications like image-text understanding and molecule-function analysis~\cite{li2023blip, cao2024presto, liu2023molca}.


The advances in MLLMs have led to significant developments in text-based protein understanding~\citep{liu-etal-2024-prott3, proteindt, evolla, lv2024prollama, liu2024evollama}. The input of most tasks generally consists of a protein sequence paired with natural language text, while the output represents a functional description.
Given that proteins can be represented as amino acid sequences, they are naturally compatible with LLMs and can be processed in two primary ways (Fig.~\ref{fig: teaser}a): (1) directly as textual inputs to a language model in a decoder-only or encoder-decoder architecture~\citep{fang2023mol, lv2024prollama, biogpt, BioT5}, or (2) as an external modality, where specialized encoders first extract high-quality protein representations before alignment with LLMs for downstream tasks~\citep{liu2024evollama}. 
To evaluate such protein-text multimodal models, several benchmarks have been introduced, covering key tasks such as protein function prediction, subcellular localization, catalytic activity, and protein design~\citep{fang2023mol}. 
The model’s performance is assessed by comparing the model's outputs against ground-truth annotations.
While promising, existing methods raise key questions:
\begin{tcolorbox}[
    colback=blue!10!white,  
    colframe=blue!75!black, 
    rounded corners,        
    boxrule=0.5mm,          
    parbox=false            
]
\small
\textbf{Q1:} Can LLMs truly understand protein sequences?

\vspace{1mm}

\noindent \textbf{Q2:} Are current benchmarks suitable for protein understanding tasks?
\end{tcolorbox}


To answer the two questions, we recall that retrieval methods have long served as fundamental approaches in protein tasks, leveraging sequence alignment and database search techniques to identify functional and structural similarities~\citep{lee2007predicting, higdon2010modeling, eswar2006comparative}. These well-established methods provide a natural baseline for evaluating whether modern LLMs offer genuine advances in protein understanding or merely replicate retrieval paradigms through alternative mechanisms.
We therefore tackle Q1 by first comparing traditional retrieval methods against LLMs.
Surprisingly, our analysis reveals that \textit{simple retrieval-based approaches} can match or even outperform current LLMs in protein sequence understanding, challenging the prevailing view that LLMs are inherently superior in this domain.

Through a comprehensive analysis of prevailing protein-text datasets and evaluation metrics, we identify \textit{two key limitations} in current benchmarks: (1) significant data leakage issues that compromise benchmark validity, and (2) metrics that fail to adequately capture model performance on biologically meaningful tasks. We systematically evaluate both LLM-based and retrieval-based approaches across existing datasets, revealing that MLLMs primarily generate outputs by memorizing and reproducing similar input features. Motivated by our analysis and findings, we propose a more rigorous benchmark for text-based protein understanding and introduce an efficient protein knowledge retrieval system, which achieves the state-of-the-art performance in protein understanding by Retrieval-Augmented Protein Modeling (RAPM).

\noindent Our contributions are summarized as follows:
\begin{itemize}[leftmargin=10pt, noitemsep]
  \item We evaluate existing protein-text benchmarks, revealing data leakage and metric limitations, and propose the new Prot-Inst-OOD dataset and Bio-Entity BLEU metric.
  \item We systematically compare fine-tuned LLMs with retrieval-based methods, demonstrating that fine-tuning is unnecessary for specific tasks.
  \item We propose RAPM, a Retrieval-Augmented Protein Modeling framework with a dual-indexed protein knowledge database for enhancing LLM in protein understanding tasks.
\end{itemize}



\section{Related Works}

\begin{figure*}[!htp]
    \centering
    \includegraphics[width=.96\textwidth]{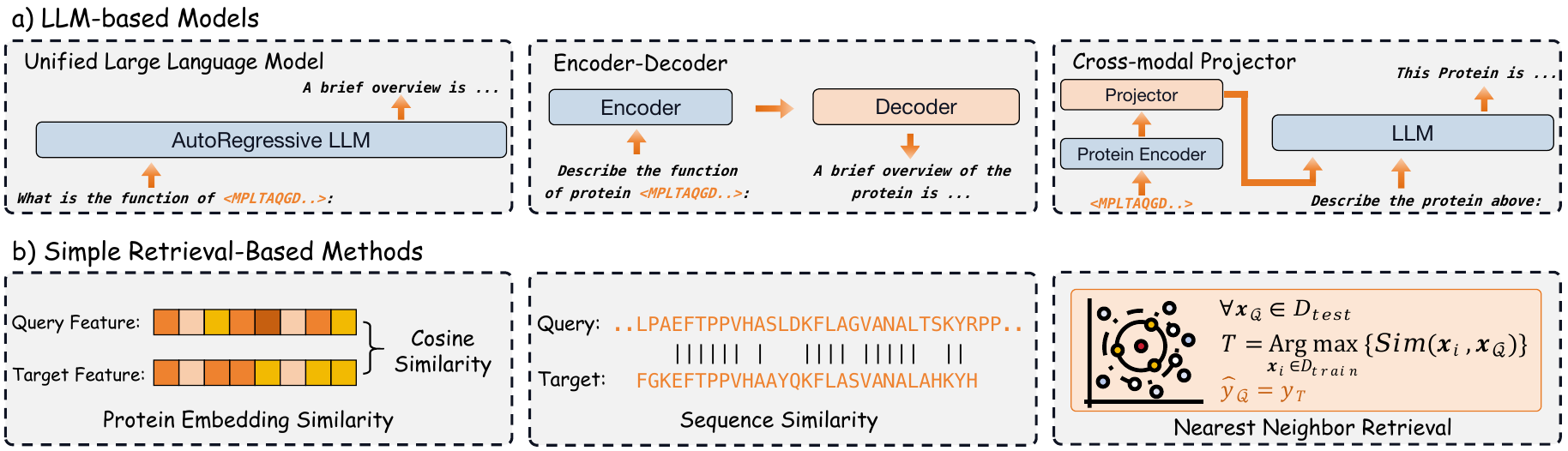}
    \vspace{-0.2cm}
    \caption{a) Three typical LLM-based approaches for text-based protein understanding. b) Simple nearest-neighbor based retrieval with protein embedding or sequence similarities.}
    \label{fig:baseline}
    \vspace{-0.5cm}
\end{figure*}

This section provides an overview of prior research focused on three key aspects: (1)  applications of language models to protein science, (2) existing benchmarks for protein understanding, and (3) retrieval-based approaches in protein research.

\subsection{Language Models in Protein}
\label{sec: related PLMs}

Protein language models (PLMs) have successfully adapted Transformer architectures to represent protein sequences as biological tokens, enabling advances in protein embedding~\cite{doi:10.1126/science.ads0018, brandes2022proteinbert, elnaggar2021prottrans, cao2021tale, ProtGO, chen2024xtrimopglm, chen2024unifying, xue2022multimodal} and design~\cite{madani2023large, nijkamp2023progen2, lv2024prollama, ferruz2022protgpt2}. However, their inability to integrate textual information limits cross-modal reasoning, prompting recent work to develop mixed protein-text models. These approaches include \textit{Contrastive Learning Methods}~\cite{protst, wu2024proteinclip} that align protein sequences with text, \textit{Bioknowledge-Augmented Pre-training}~\cite{ferruz2022protgpt2, taylor2022galactica, lv2024prollama, BioT5, zhuo2024protllm, liu2024evollama} that leverage large protein-text corpora, and \textit{Multi-modal LLMs}~\cite{liu-etal-2024-prott3, abdine2024prot2text, wang2024protchatgpt, chen2024unifying, ma2025prottex, xiang2024fapm} that project protein embeddings into LLM spaces. Despite progress, scaling these methods to larger LLMs remains challenging due to prohibitive retraining costs and catastrophic forgetting~\cite{kirkpatrick2017overcoming}, motivating research into parameter-efficient adaptation strategies.

\subsection{Related Benchmarks}
\label{sec: related benchmark}
To advance research on protein-text hybrid models, several relevant benchmarks have been proposed. These benchmarks can be categorized into two types: (1)~\textit{Protein Captioning Tasks}, where only the protein sequence is input and a corresponding textual description is generated (e.g., the Swiss-Prot~\cite{bairoch2000swiss} and ProteinKG datasets~\cite{zhang2022ontoprotein}), and (2)~\textit{Protein Question-Answering Tasks}, where both a protein sequence and a question are provided as input, and the model must generate an answer based on the protein and the query (e.g., Mol-Instructions~\cite{fang2023mol}, UniProtQA~\cite{luo2024biomedgpt}, ProteinLMBench~\cite{shen2024fine}, Prot2Text~\cite{abdine2024prot2text}). To evaluate model performance on these benchmarks, researchers typically employ standard NLP metrics such as ROUGE~\cite{lin2004rouge}, BLEU~\cite{papineni2002bleu}, and METEOR~\cite{banerjee2005meteor} to measure the similarity between predicted answers and ground-truth references. Models that perform well on these tasks can be applied to automated protein annotation, protein design, and protein property-related QA, thereby facilitating progress in the field.




\subsection{Protein Related Retrieval-Based Methods}
Retrieval-based approaches are fundamental to protein science, grounded in the well-established biological principle that sequence homology implies evolutionary conservation and functional similarity~\cite{pearson2013introduction}. Single-sequence alignment approaches \cite{altschul1990basic,buchfink2015fast,steinegger2017mmseqs2,van2022foldseek} and multiple sequence alignment tools~\cite{remmert2012hhblits,johnson2010hidden} are extensively used in bioinformatics for identifying highly homologous sequences. Many protein models utilize retrieval methods to assist downstream tasks, where AlphaFold2~\cite{jumper2021highly}, MSA-Transformer~\cite{rao2021msa}, and RosettaFold~\cite{baek2021accurate} employ multiple sequence alignment results to aid property prediction or structure folding. Furthermore, retrieval-based approaches~\cite{tan2024retrieval,shaw2024protex,sgarbossa2025rag,jin2024prollm,li2024retrieval,ma-etal-2024-retrieved} have demonstrated the feasibility of using retrieval tools to enhance LLM-based predictions in protein research.

\section{Analysis}


Despite the broad usage of LLMs for protein understanding tasks, it remains unclear whether they truly understand protein sequences or simply memorize patterns. To answer this question, we conduct a systematic comparison between LLMs and retrieval-based methods, analyzing their performance and studying what LLMs actually learn.

\begin{table*}[ht]
    \small
    \centering
    \renewcommand{\arraystretch}{0.8}
    \resizebox{\textwidth}{!}{ 
    \begin{tabular}{l|c|c|cccc|c}
    \toprule
    \multirow{2}{*}{\textbf{Model}} & 
    \multirow{2}{*}{\textbf{Arch.}} & 
    \multirow{2}{*}{\textbf{SFT}} &
    \multicolumn{4}{c|}{\textbf{Mol-Instructions/Protein (ROUGE-L)}} &
    \multirow{2}{*}{\textbf{Avg.}} \\
    \cmidrule(lr){4-7}
     & & & \textbf{Function} & \textbf{Description} & \textbf{Domain} & \textbf{Catalytic} & \\
    \midrule
    Galactica-1.3B\textit{-SFT} & Decoder-only & $\checkmark$ & 7.1  & 48.2 & \underline{55.3} & 30.2 & 35.2 \\
    BioGPT-347M\textit{-SFT} & Decoder-only & $\checkmark$ & 50.9 & 49.7 & \textbf{55.4} & 54.2 & 52.5 \\
    ProLLaMA-7B\textit{-SFT} & Decoder-only & $\checkmark$ & 48.6 & 20.3 & 46.7 & 39.3 & 38.7 \\
    Mol-Instructions-7B & Decoder-only & $\checkmark$ & 43.0 & 44.0 & 46.0 & 52.0 & 46.2 \\
    Llama-3.1-8B\textit{-SFT} & Decoder-only & $\checkmark$ & 52.1 & 54.2 & 51.2 & 59.6 & 54.2 \\ 
    BioT5-Plus-252M & Encoder-Decoder & $\checkmark$ & 56.6 & 68.0 & 53.4 & 71.8 & 62.4 \\
    EvoLLaMA-8.8B & MLP-Projector & $\checkmark$ & 48.0 & 50.0 & 50.0 & 60.0 & 52.0 \\
    FAPM-10B & Q-Former & $\checkmark$ & \textbf{60.9} & 64.0 & 52.7 & \textbf{76.0} & 63.4 \\
    \midrule
    \rowcolor{gray!20}
    MMSeqs2-Align & Retrieval & $\times$ & \underline{60.2} & \textbf{76.0} & 55.2 & \underline{75.6} & \textbf{66.7} \\
    \rowcolor{gray!20}
    ESM2-Embedding & Retrieval & $\times$ & 59.7 & \underline{74.9} & 54.5 & 75.2 & \underline{66.0} \\
    \bottomrule
    \toprule
    \multirow{2}{*}{\textbf{Model}} & 
    \multirow{2}{*}{\textbf{Arch.}} & 
    \multirow{2}{*}{\textbf{SFT}} &
    \multicolumn{4}{c|}{\textbf{UniProtQA Benchmark}} &
    \multirow{2}{*}{\textbf{Avg.}} \\
    \cmidrule(lr){4-7}
     & & & \textbf{BLEU-2} & \textbf{BLEU-4} & \textbf{ROUGE-L} & \textbf{METEOR} & \\
    \midrule
    Llama2-7B-Chat & Decoder-only & $\times$ & 1.9 & 2.0 & 0.9 & 5.2 & 2.5 \\
    Llama2-7B-SFT & Decoder-only & \checkmark & 34.4 & 31.3 &  59.3 & 70.7 & 48.9 \\
    BioMedGPT-10B  & Q-Former & \checkmark & {57.1} & {53.5} & {62.2} & {75.4} & 62.0 \\
    \midrule
    \rowcolor{gray!20}
    MMSeqs2-Align  & Retrieval & $\times$  & \textbf{85.5} & \textbf{84.2}  & \textbf{91.4} & \textbf{91.7} & \textbf{88.2} \\
    \bottomrule
    \toprule
    \multirow{2}{*}{\textbf{Model}} & 
    \multirow{2}{*}{\textbf{Arch.}} & 
    \multirow{2}{*}{\textbf{SFT}} &
    \multicolumn{2}{c}{\textbf{Swiss-Prot}} & \multicolumn{2}{c|}{\textbf{ProteinKG25}} & 
    \multirow{2}{*}{\textbf{Avg.}} \\
    \cmidrule(lr){4-5}  \cmidrule(lr){6-7}
     & & & \textbf{BLEU-2} & \textbf{ROUGE-L} & \textbf{BLEU-2} & \textbf{ROUGE-L} & \\
    \midrule
    Galactica-1.3B\textit{-SFT} & Decoder-only & \checkmark & {42.4} & {42.4} & {64.9} & {62.5} & 52.9 \\
    ProtT3  & Q-Former & \checkmark & {55.0} & {62.1} & {76.5} & {71.4} & 66.2 \\
    \midrule
    \rowcolor{gray!20}
    MMSeqs2-Align  & Retrieval & $\times$  & \textbf{75.7} & \textbf{80.6}  & \textbf{80.8} & \textbf{76.2} & \textbf{78.3} \\
    \bottomrule
    
\end{tabular}
    }
    \vspace{-0.1cm}
    \caption{Performance comparison of LLM-based and retrieval-based methods across two text-based protein understanding benchmarks. \textbf{Arch.} denotes the model architecture. \textbf{SFT} indicates whether the model has undergone supervised fine-tuning on the training set. \textbf{Bold} denotes the best. \underline{Underline} denotes the second best. }
    \label{tab:benchmark-molinst}
    \vspace{-0.4cm}
\end{table*}

\subsection{Retrieval vs. LLM in Existing Tasks}
\label{sec: leakage}
We first evaluate both LLM-based and retrieval-based approaches on existing benchmarks, with the following experimental setup:
\begin{itemize}[noitemsep, leftmargin=8pt]
    \item \textbf{LLM-based approach} (Fig.~\ref{fig:baseline}a): After fine-tuning the model on the training dataset, we processed test samples using next-token prediction to generate answers.
    \item \textbf{Retrieval-based approach} (Fig.~\ref{fig:baseline}b): For each test sample, we retrieve the most similar protein sequence from the training set and use its annotation as the answer.
\end{itemize}

For LLM-based methods, we test a variety of model architectures, including 5 unimodal LLMs (Galactica-1.3B-SFT~\cite{taylor2022galactica}, BioGPT-347M-SFT~\cite{biogpt}, ProLLaMA-7B-SFT~\cite{lv2024prollama}, Mol-Instructions-7B~\cite{fang2023mol}, and Llama-3.1-8B-SFT~\cite{llama3}), 1 encoder-decoder model (BioT5-Plus~\cite{BioT5}), and 2 multi-modal LLMs (EvoLLAMA-8.8B~\cite{liu2024evollama} and FAPM-10B~\cite{xiang2024fapm}). For retrieval approaches, we employ MMSeqs2~\cite{steinegger2017mmseqs2} for sequence retrieval and ESM-2-650M~\cite{esm2} as the protein sequence encoder for embedding similarity.
Our evaluation results (Table \ref{tab:benchmark-molinst}) highlight a key finding: \textbf{all current deep learning methods underperform retrieval-based approaches on these benchmarks}.
We find that multi-modal LLMs merely match the performance of retrieval methods, while unimodal LLMs demonstrate poorer results. More critically, fine-tuning LLMs requires significant GPU resources, whereas ESM2-based retrieval only needs to compute protein embeddings, and MMSeqs2 retrieval completes 100 million comparisons within 1 minute using only one CPU.

\begin{figure}[!htp]
    \centering
    \includegraphics[width=0.93\linewidth]{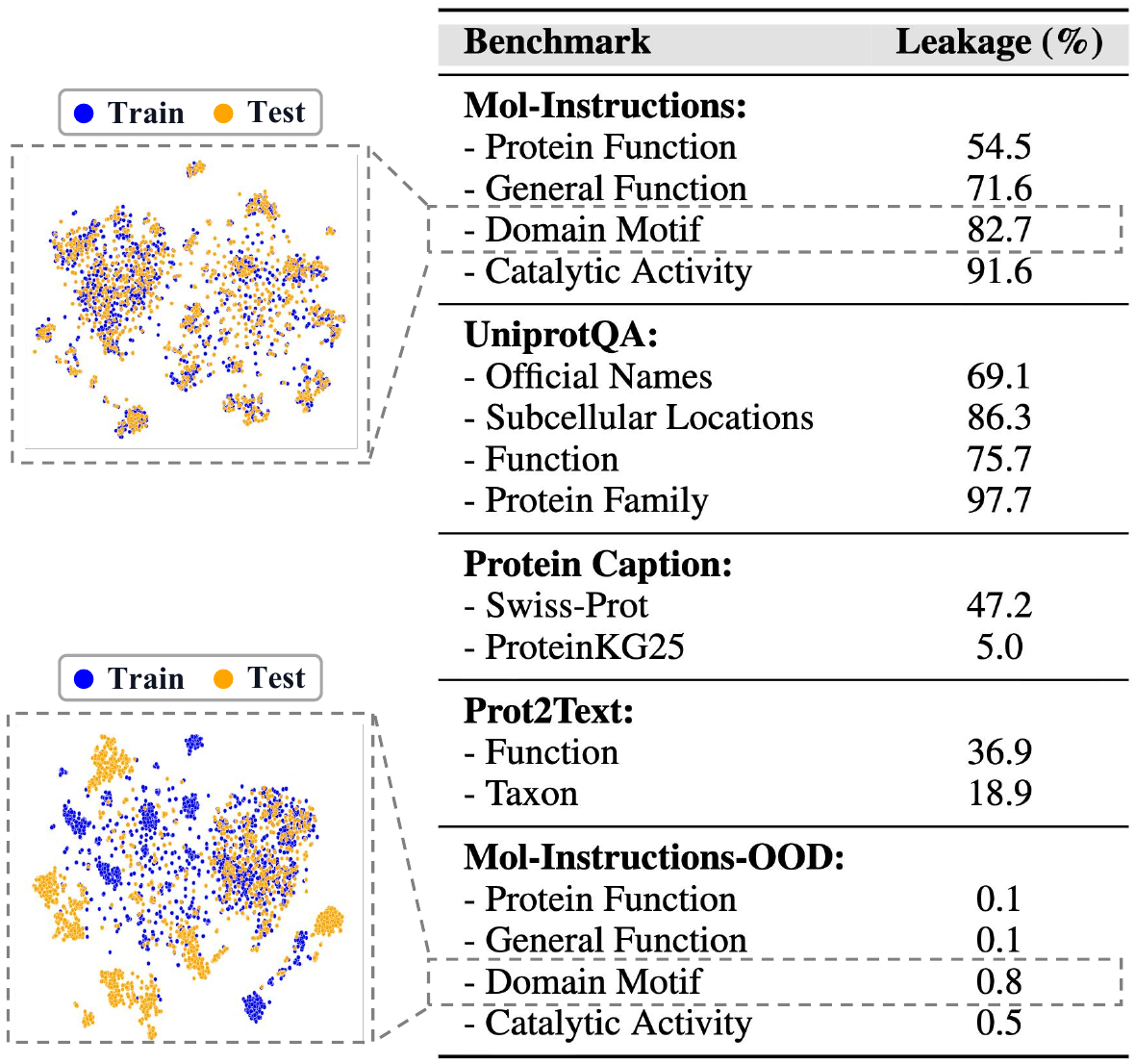}
    \caption{
    We evaluate the degree of data leakage in both existing benchmarks and OOD benchmarks. ``Leakage" is defined as the probability that test set samples can directly retrieve similar samples with the same label from the training set.
    }
    \label{fig:data_leakage}
    \vspace{-0.4cm}
\end{figure}

To investigate why the retrieval-based methods beat LLMs, we first examine the data distribution in current benchmarks. The t-SNE visualization of the ESM2 embeddings of the proteins reveals samples forming distinct clusters with significant training-test contamination (Figure~\ref{fig:data_leakage}). \textbf{Prot2Text} is the multimodal dataset collected by \citeauthor{abdine2024prot2text}, where they reduced the overlap between the training and test samples.
We then quantify the level of label leakage by the percentage of test samples whose label can be directly obtained by retrieving the most similar sample (right table in Figure~\ref{fig:data_leakage}). It is easy to see that the leakage rates exceed 50\% for most tasks, surpassing 95\% in some extreme cases. The process of protein function annotation possibly causes such pervasive label leakage and suggests that models fine-tuned on these benchmarks predominantly memorize dataset-specific features rather than develop meaningful biological understanding.

\subsection{What do LLMs Learn?}
\label{sec:what_do_llm_learn}
A fundamental question in text-based protein understanding is whether LLMs genuinely comprehend protein knowledge or simply act as sophisticated pattern matches based on input similarities.
To address this, we perform a fine-grained comparative analysis between LLM-based and retrieval-based approaches. Specifically, we visualize and compare their performance across all test samples in the Protein Function task. This analysis enables us to distinguish whether LLMs predict properties based on protein sequence features or simply learn to replicate labels from similar training samples.

\begin{figure}[!ht]
    \centering
    \includegraphics[width=0.8\linewidth]{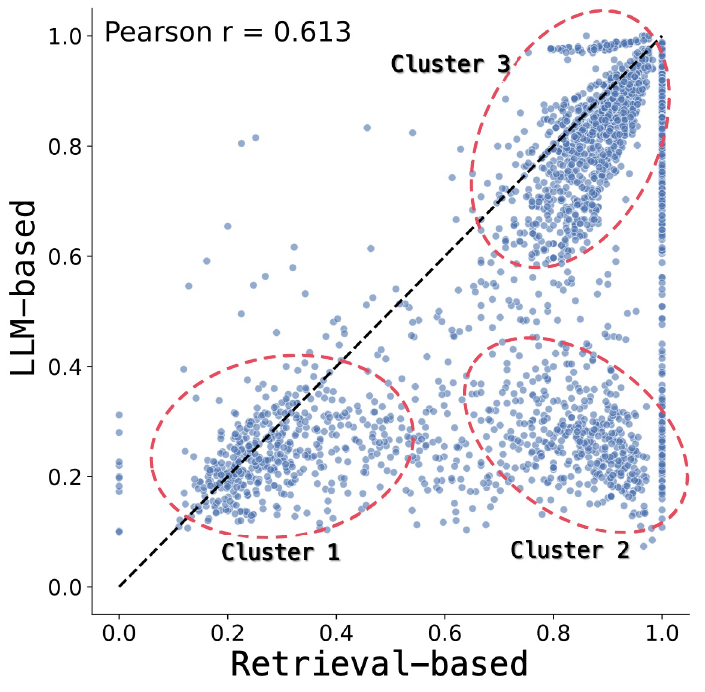}
    \vspace{-0.2cm}
    \caption{
    The ROUGE-L score distributions of retrieval-based methods versus LLM-based methods for all test samples in the General Function task.
    }
    \label{fig: what_mllm_learn}
    \vspace{-0.3cm}
\end{figure}

We compare the performance of the retrieval method to that of the LLM method under the measurement of ROUGE-L (Fig.~\ref{fig: what_mllm_learn}). The majority of samples fall below the $y=x$ reference line and naturally separate into three clusters: 
\vspace{-0.2cm}
\begin{itemize}[noitemsep, leftmargin=8pt]
    \item Cluster 1: Both retrieval and LLM methods fail to predict the protein function.
    \item Cluster 2: The retrieval method correctly predicts the function while the LLM method fails
    \item Cluster 3: Both methods demonstrate competent performance.
\end{itemize}

\vspace{-0.4cm}

\noindent Our analysis demonstrates that LLMs fine-tuned for protein function prediction fail to surpass the performance of retrieval-based approaches for most test samples, indicating they primarily serve as a less effective substitute for retrieval approaches.

\subsection{Is Retrieval a Silver Bullet?}
\noindent We further investigate if traditional retrieval methods are a silver bullet as shown above. Using the entire training set as retrieval candidates, rather than creating separate pools for each subtask, leads to significant performance degradation for different methods (Table~\ref{tab:retrieve_bad}). Such task-specific pools are impractical in practice, given task diversity and continuous change. Furthermore, traditional methods often return only the top-1 match, preventing multi-source aggregation and lacking flexibility.
\begin{table}[!htbp]
\centering
\small
\renewcommand{\arraystretch}{0.9}
\setlength{\tabcolsep}{3pt} 
\resizebox{\linewidth}{!}{
\begin{tabular}{lcccc}
\toprule
\rowcolor{gray!20}
\textbf{Retrieval}   & \textbf{Function} & \textbf{Description} & \textbf{Domain} & \textbf{Catalytic} \\
\midrule
MMSeqs2 & 60.2 &76.0 &55.2 &75.6 \\
MMSeqs2$_\text{all}$ &40.0(\textcolor{red}{$\downarrow$34\%}) &25.4(\textcolor{red}{$\downarrow$67\%}) &36.7(\textcolor{red}{$\downarrow$34\%}) &37.6(\textcolor{red}{$\downarrow$50\%}) \\
\midrule
ESM2-Embed &59.7 &74.9 &54.5 &75.2 \\
ESM2-Embed$_\text{all}$ &38.7(\textcolor{red}{$\downarrow$35\%}) &17.6(\textcolor{red}{$\downarrow$77\%}) &36.7(\textcolor{red}{$\downarrow$33\%}) &26.8(\textcolor{red}{$\downarrow$64\%}) \\
\bottomrule
\end{tabular}
}
\vspace{-0.2cm}
\caption{Performance degradation of retrieval methods with the full corpus as the candidate pool.}
\label{tab:retrieve_bad}
\vspace{-0.3cm}
\end{table}



\noindent \textbf{Summary:} For practical usage, neither retrieval-based methods nor LLMs provide satisfactory protein understanding, which suggests a hybrid framework that synergistically combines the precision of retrieval with the reasoning capacity of LLMs. 

\section{Methods: Combine Retrieval \& LLM and New Benchmark} 

\begin{figure*}[!htp]
    \centering
    \includegraphics[width=.97\textwidth]{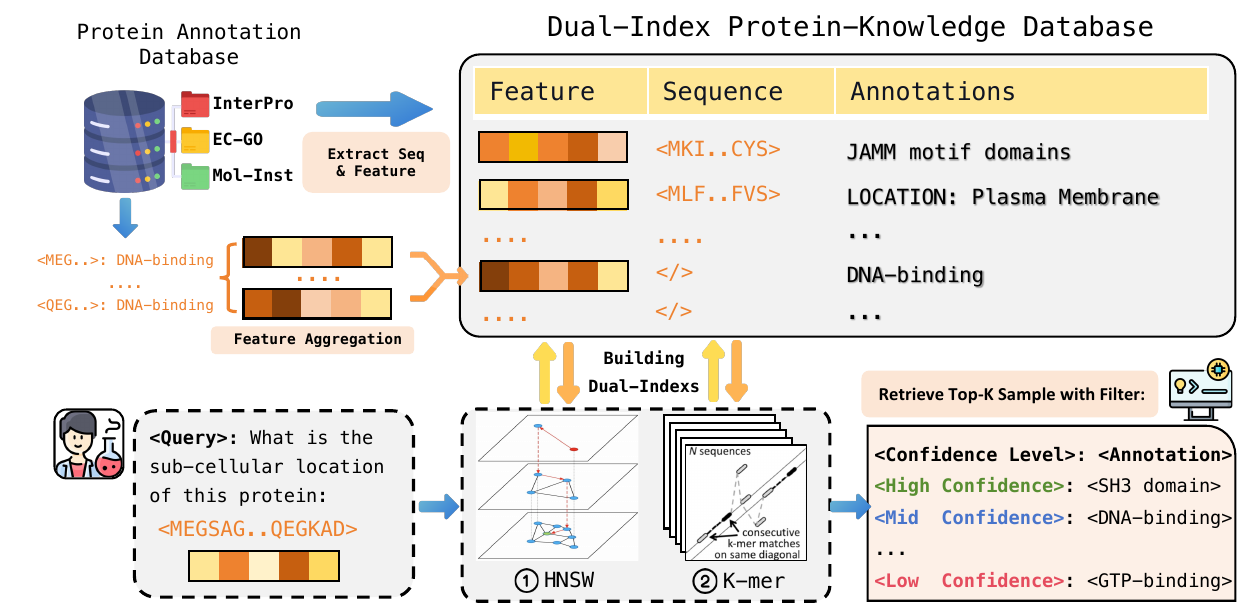}
    \vspace{-0.2cm}
    \caption{
        We collect protein-annotation pairs from existing protein annotation databases for the Protein-Knowledge Database construction. We extract dense features of proteins using a Protein Encoder and build database indices using two indexing methods. For entries sharing identical labels, we incorporate meta-features into the database. For downstream queries, we combine scores from both indices to retrieve the Top-K relevant entities, then construct retrieval-augmented prompts after quantizing sequence similarity into \textcolor{green!50!black}{\texttt{\textbf{High}}}, \textcolor{blue!70!black}{\texttt{\textbf{Mid}}}, and \textcolor{red!80!black}{\texttt{\textbf{Low}}} confidence levels.
    }
    \label{fig:pipeline}
    \vspace{-0.3cm}
\end{figure*}



To address this need, we develop Retrieval-Augmented Protein Modeling (RAPM) based on the Retrieval-Augmented Generation (RAG) paradigm. RAG is a proven approach for enhancing LLM factual accuracy and domain knowledge~\cite{lewis2020retrieval}. Our method leverages a Bio-Knowledge Database and contextual prompts to provide LLMs with explicit protein evidence during inference, thereby improving their understanding of biological information and addressing the memorization vs. reasoning tradeoff inherent in this problem.

\subsection{Protein Knowledge Database Construction}

For optimal performance, an accurate and efficient domain retrieval system relies on a carefully curated protein knowledge database. In our database, existing biological annotations are standardized into structured \texttt{[Protein, Annotation]} tuples, indexed by amino acid sequence and embeddings. 
The details of the database construction are shown in Fig.~\ref{fig:pipeline}, including:

\begin{itemize}[noitemsep, leftmargin=8pt]
\item \textbf{Protein Annotation Data Collection.} We gather protein-annotation data from InterPro~\cite{InterPro}, EC-GO~\cite{gligorijevic2021structure}, and Mol-Instructions~\cite{fang2023mol}, extracting biological entity annotations using the method in Sec.~\ref {sec: new metric}.

\item \textbf{Dual-key Indexing.} 
We build database indices with: (1) Sequence-based Indexing with inverted K-mer indices for heuristic retrieval, and (2) Feature-based Indexing using ESM-2 extracted protein features with the HNSW algorithm for efficient indexing.

\item \textbf{Feature Aggregation.} 
For an annotation shared by multiple proteins, we compute the mean-pooled embedding of all proteins with that annotation. This aggregated feature ensures retrieval breadth while maintaining biological relevance.
\end{itemize}

\subsection{Retrieval-Augmented Protein Modeling}
Using a RAG-based approach, we provide LLMs with explicit protein evidence during inference for downstream queries.
The overall retrieval and query pipeline is illustrated in Fig.~\ref{fig:pipeline}. A critical component is our reformatted protein knowledge database, constructed by reorganizing existing protein annotations into standardized \texttt{[Protein, Annotation]} tuples. For precise and efficient retrieval, each entry is indexed using a dual-key mechanism, incorporating both amino acid sequence-based indexing with inverted K-mer indices for heuristic retrieval and feature-based indexing using ESM-2 embeddings with the HNSW algorithm for efficient similarity search (details about the indexing refer to Appendix~\ref{abla:rag_settings}). To further improve retrieval breadth, especially for annotation labels shared by multiple proteins, we aggregate the features by computing the mean-pooled embedding of all proteins associated with a common annotation and indexing these aggregated features.

Formally, given a protein query $\mathcal{Q}$, we retrieve $K$ support data points $\{d_i\}_{i=1}^K$ from this structured database by ranking candidate entries based on a similarity score $s_i$. This score is a weighted combination of sequence and embedding similarity: $s_i = \alpha \cdot \mathrm{sim}_{seq}(s, s_i) + (1-\alpha)\cdot \mathrm{sim}_{emb}(\mathbf{e}, \mathbf{e}_i)$, where $s$ and $s_i$ are protein sequences, $\mathbf{e}$ and $\mathbf{e}_i$ are the corresponding ESM-2 embeddings, and $\alpha$ is a weight parameter, currently set to $0.5$. Instead of including full sequences in the prompt, each of the top-$K$ retrieved items is formatted as a concise \texttt{[Confidence, Annotation]} tuple. The Confidence level is derived from $s_i$ based on quantiles: $> 90\%$ as \textcolor{green!50!black}{\texttt{\textbf{High }}}, $90\% > s_i > 60\%$ as \textcolor{blue!70!black}{\texttt{\textbf{Medium}}}, and $\le 60\%$ as \textcolor{red!80!black}{\texttt{\textbf{Low}}}.
The final input prompt $\mathcal{P}$ for the LLM is constructed by concatenating the query, few-shot examples, and the formatted retrieved items: $\mathcal{P} = \mathcal{Q} \oplus \mathcal{E}_{\text{few-shot}} \oplus \mathcal{R}_{1:k}$, where $\mathcal{R}_{1:k}$ represents the formatted top-$K$ entries and $\mathcal{E}_{\text{few-shot}}$ are demonstrations from the training dataset included to help the LLM understand the task format and reasoning. The LLM is then conditioned on $\mathcal{P}$ to predict the answer: $\hat{y} = \mathrm{LLM}(\mathcal{P})$.

\subsection{Novel Benchmark Proposal}
\label{sec:benchmark}
Existing benchmarks (Sec~\ref{sec: related benchmark}) rely on NLP-derived metrics like token or sentence similarity, implicitly assuming equal importance for all answer components. This approach is fundamentally flawed for biological QA tasks. In protein-related questions, responses frequently include standardized template structures while the critical biological information is concentrated in just a few content words, which typically appear in the final portion of the answer. Consider the following example:

\begin{tcolorbox}[colback=gray!10!white, boxrule=0pt, sharp corners=south]
\footnotesize
\textbf{Ground Truth:}\\
\texttt{Upon evaluating your submitted sequence, our predictive algorithms suggest the presence of:  \underline{ABC transporter domains}}
\vspace{0.1cm}
\\
\textbf{Prediction 1 (True Answer):}\\
\texttt{\textcolor{red}{The sequence you provided has been analyzed for potential protein domains or motifs. The results are}\textcolor{blue}{: ABC transporter domains}} 
\vspace{-0.1cm}
\begin{tcolorbox}[colback=orange!30, boxrule=0.5pt, width=.7\linewidth, left=2pt, right=2pt, top=1pt, bottom=1pt, fontupper=\footnotesize]
\texttt{ROUGE-L = 0.27; BLEU = 0.04}
\end{tcolorbox} 
\textbf{Prediction 2 (False Answer):}\\
\texttt{\textcolor{blue}{Upon evaluating your submitted sequence, our predictive algorithms suggest the presence of: }\textcolor{red}{GGDEF, MHYT, EAL domains}}
\vspace{-0.1cm}
\begin{tcolorbox}[colback=orange!30, boxrule=0.5pt, width=.7\linewidth, left=2pt, right=2pt, top=1pt, bottom=1pt, fontupper=\footnotesize]
\texttt{ROUGE-L = 0.83; BLEU = 0.73}
\end{tcolorbox} 
\vspace{-2mm}
\end{tcolorbox}
\vspace{-3mm}
\begin{center}
\footnotesize
\textcolor{blue}{\textbf{Blue}}: Matched Part \quad \textcolor{red}{\textbf{Red}}: Mismatched Part
\end{center}

\noindent Although \textbf{Prediction 2} achieves much higher NLP metric scores, its information is biologically inaccurate. This discrepancy highlights a critical flaw of current evaluation metrics: they prioritize superficial text overlap, particularly in generic template segments, and are insensitive to errors in the core biological content.

To address the data leakage in Sec~\ref{sec: leakage} and metric validity issues identified above, we construct a new protein domain benchmark with novel task partitions to avoid data leakage and a BLEU-like metric specifically designed for biological entities.
 
\paragraph{Data Unification and Clustering.}
To address data leakage, we reconstruct protein-text datasets by integrating four protein understanding tasks in Mol-Instructions \cite{fang2023mol} and captioning tasks in Swiss-Prot \cite{bairoch2000swiss}. For captioning tasks, we generate instructions from original annotations, forming a unified dataset (Pro-Inst-OOD). The OOD construction involves two steps: (1) \textit{Low similarity (Low-Sim) split}: Based on MMSeqs2 clustering with an 8:2 class split, mitigating general leakage. (2) \textit{Out-of-Distribution (OOD)}: Filters Low-Sim split by removing test samples for which answers can be retrieved from the training set, preventing reliance on retrieval. 
This creates the Pro-Inst-OOD benchmark (construction details in Appendix \ref{app:dataset}).

\paragraph{Metric Design for Biological QA.}
\label{sec: new metric}
Existing NLP metrics like ROUGE and BLEU are inadequate for biological QA, failing to capture biological nuances such as order-invariant entity lists by treating all tokens equally. To address this, we propose \textbf{Entity-BLEU}, a biological entity-focused metric analogous to BLEU. It works by first extracting biological entities from predictions and references using a knowledge base derived from databases like InterPro, EC-GO, and Mol-Instructions labels. A detailed biological entities can be seen in Appendix~\ref{abla:entity_bleu}. The standard BLEU score is then computed on these extracted entity sequences. Formally, Entity-BLEU is given by: 
\begin{equation} 
\medmath{\text{Entity-BLEU} = \text{BP} \cdot \exp\left(\sum_{n=1}^{N} w_n \log p_n\right)} 
\end{equation}
where BP is the brevity penalty term, $w_n$ are the weights for n-gram precision scores $p_n$ (typically $N\in\{1,2,3,4\}$), and all calculations are performed on the extracted Bio-Entity sequences rather than raw text.

\begin{table*}[!htbp]
    \centering
    \small
    \setlength{\tabcolsep}{3pt}
    \renewcommand{\arraystretch}{0.94}
    \resizebox{0.95\textwidth}{!}{
    \begin{tabular}{lcccccccccc}
    \toprule
   \multirow{2}{*}{\textbf{Model}} & \multirow{2}{*}{\makecell{\textbf{\#Train}\\ \textbf{Params}}} & \multirow{2}{*}{\textbf{Retrieval}} & \multicolumn{2}{c}{\textbf{Function}} & \multicolumn{2}{c}{\textbf{Description}} & \multicolumn{2}{c}{\textbf{Domain}} & \multicolumn{2}{c}{\textbf{Catalytic}} \\
    \cmidrule(lr){4-5} \cmidrule(lr){6-7} \cmidrule(lr){8-9} \cmidrule(lr){10-11}
    & & & \textbf{E-BLEU} & \textbf{RG-L} & \textbf{E-BLEU} & \textbf{RG-L} & \textbf{E-BLEU} & \textbf{RG-L} & \textbf{E-BLEU} & \textbf{RG-L} \\
    \midrule
    \rowcolor{gray!20}
    \multicolumn{11}{l}{\textbf{Fine-tuned LLM}} \\
    BioT5+                & 252M & None & 3.7 & 36.3 & 0.1 & 30.4 & 0.1 & 37.9 & 0.2 & 39.6 \\
    Llama-3.2-1B       & 1.0B & None & 16.2 & \textbf{43.4} & 4.4 & \textbf{43.0} & 3.9 & 43.0 & 1.8 & 44.0 \\
    BioGPT   & 347M & None & 6.8 & 41.6 & 0.3 & \underline{34.3} & 1.1 & \underline{44.3} & 0.6 & 43.9 \\
    Galactica-1.3B        & 1.3B & None & 14.4 & \underline{43.2} & 1.2 & 32.6 & 3.8 & \textbf{45.3} & 0.9 & 43.3 \\
    ProLLama-7B*                & 19M & None & 6.3 & 39.8 & 0.5 & 30.2 & 1.0 & 41.7 & 0.4 & 41.8 \\
    \midrule
    \rowcolor{gray!20}
    \multicolumn{11}{l}{\textbf{Retrieval-based}} \\
    MMSeqs2               & N/A & Seq & 11.9 & 28.5 & 3.7 & 25.8 & 2.0 & 16.3 & 2.6 & 21.8 \\
    ESM-2-650M        & 650M & Emb. & 10.8 & 29.6 & 3.8 & 26.5 & 2.1 & 17.5 & 2.8 & 23.1 \\
    \midrule
    \rowcolor{gray!20}
    \multicolumn{11}{l}{\textbf{Task-prompted LLM}} \\
    Llama-3.3-70B$_{\text{w/ Few-shot}}$  & N/A & None & 0.3 & 29.5 & 1.0 & 27.1 & 0.1 & 36.8 & 0.1 & 44.5 \\
    DeepSeek-V3$_{\text{w/ Few-shot}}$         & N/A & None & 0.2 & 28.3 & 0.3 & 25.5 & 0.0 & 35.3 & 0.9 & 19.6 \\
    GPT-4.1$_{\text{w/ Few-shot}}$        & N/A & None & 0.1 & 31.9 & 0.2 & 26.1 & 0.1 & 38.7 & 0.1 & 40.8 \\
    \midrule
    \rowcolor{gray!20}
    \multicolumn{11}{l}{\textbf{RAPM-based}} \\
    Llama-3.3-70B$_{\text{w/ RAPM}}$  & N/A & Seq+Emb. & \underline{41.5} & 37.5 & \underline{16.9} & 25.4 & 7.3 & 11.1 & \underline{23.5} & \underline{44.6} \\
    DeepSeek-V3$_{\text{w/ RAPM}}$  & N/A & Seq+Emb. & 35.3 & 31.2 & 13.8 & 24.4 & \underline{8.8} & 17.9 & 16.3 & 21.0 \\
    GPT-4.1$_{\text{w/ RAPM}}$         & N/A & Seq+Emb. & \textbf{46.6} & 27.4 & \textbf{20.9} & 30.1 & \textbf{32.0} & 22.5 & \textbf{38.9} & \textbf{46.4} \\
    \bottomrule
\end{tabular}}
    \vspace{-0.2cm}
    \caption{Performance of different approaches in Prot-Inst-OOD, each evaluated with E-BLEU(Entity-BLEU) and RG-L(ROUGE-L). "*" means using LoRA \cite{hu2021lora} fine-tuning. \textbf{Bold} for best, \underline{underline} for second best.}
    \label{tab:main_results_reorg}
    \vspace{-0.3cm}
\end{table*}

\section{Experiments}
This section comprehensively evaluates the proposed RAPM with the benchmark and metric proposed in Sec.~\ref{sec:benchmark}. Experimental results demonstrate that retrieval-based methods show significant performance degradation in the Pro-Inst-OOD benchmark.
In ablation studies, we assess the impact of the number of items retrieved, database indexing methods, and prompt construction approaches for retrieval.


\subsection{Experimental Setup}
We evaluate four representative methodological approaches under our novel dataset splitting strategy described in Sec.~\ref{sec:benchmark}:

\begin{enumerate}[noitemsep, leftmargin=9pt]
\vspace{-0.1cm}
\item \textbf{Fine-tuned LLMs:} These approaches fine-tune pre-trained language models (BioGPT, BioT5+, Llama) on the training set and inference on the test set.

\item \textbf{Retrieval-based methods:}  For each test input, these approaches use the label of the retrieved most similar training sample as the predictions.

\item \textbf{Task-Prompted LLMs:} These approaches employ few-shot prompting frameworks with general-purpose LLMs (Llama-3.3, DeepSeek-V3, GPT-4.1), denoted by the subscript ``few-shot'', to generate predictions without retrieval augmentation.

\item \textbf{RAPM methods (Ours):} Our method retrieves top-K relevant samples from a protein knowledge database, constructs augmented prompts with these samples, and leverages general LLMs to generate context-aware responses.
\end{enumerate}

\vspace{-4mm}

\noindent We test all subtasks in Pro-Inst-OOD, including ``Protein Function'', ``Functional Description'', ``Domain/Motif'', ``Catalytic Activity'', and ``Protein Caption'', using the standard NLP metrics (ROUGE-L) and our proposed metric, Entity-BLEU ($N=2$). Detailed fine-tuning hyperparameters, retrieval settings, and RAPM prompt can be seen in Appendix~\ref{abla:hyper_params} and \ref{abla:rag_settings}. Note that for a fair comparison between RAPM methods and fine-tuned LLMs, we exclude all extra-training-set data during retrieval to prevent potential data leakage. In addition, all subtasks are trained and tested together, but the results are reported separately.

\subsection{Pro-Inst-OOD Performance}


Table~\ref{tab:main_results_reorg} summarizes the main results on existing benchmarks for four representative methodological approaches, and we observe the following key results:
(1) When evaluated in OOD settings, the RAPM method achieves the highest Entity-BLEU scores, outperforming retrieval-based methods and demonstrating substantial improvements over fine-tuned and task-prompted LLMs. Beyond the superior performance, RAPM requires substantially fewer computational resources than fine-tuned LLMs and demonstrates a stronger capability to handle diverse tasks than retrieval-based methods. 
(2) When comparing ROUGE-L and Entity-BLEU scores of different methods, we observe a poor correlation between them, particularly for fine-tuned LLMs, which have high ROUGE-L scores and low Entity-BLEU scores. As discussed in Sec~\ref{sec:what_do_llm_learn}, we owe this to the fact that fine-tuned LLMs primarily focus on learning irrelevant response patterns rather than understanding protein sequences. 


\subsection{Entity-BLEU Metric Analysis}
\vspace{-2mm}

\begin{table}[h]
\small
\centering
\setlength{\tabcolsep}{3pt}
\renewcommand{\arraystretch}{0.9}
\resizebox{\linewidth}{!}{
\begin{tabular}{lcccc}
\toprule
\rowcolor{gray!20}
 & \textbf{Function} & \textbf{Description} & \textbf{Domain} & \textbf{Catalytic} \\
\midrule
Entity-BLEU              & 34.32 & 24.29 & 24.80 & {27.81} \\
\midrule
Token-based F1           & 61.28 & 52.65 & 38.83 & {54.64} \\
LLM-scores               & 55.80 & 32.09 & 23.78 & {34.71} \\
\midrule
Pearson. (w/ F1) & 0.88 & 0.91 & 0.88 & 0.89 \\
Pearson. (w/ LLM)     & 0.79  & 0.73  & 0.67  & {0.73} \\
\bottomrule
\end{tabular}
}
\caption{RAPM Performance (\%) in Prot-Inst-OOD and the Pearson Correlation between Entity-BLEU and the other metrics.}
\vspace{-3mm}
\label{tab:entitybleu_multi_metric}
\end{table}

To evaluate the validity of Entity-BLEU, we compare it with two existing frequently used metrics for open-ended query answering: Token-based accuracy and LLM-as-a-Judge~\cite{gu2024survey}. For Token-based accuracy, we evaluate Token-based F1 using the formula \ref{token-based-f1}, where $\mathrm{Pred}$ and $\mathrm{GT}$ represent the bio-entities sets from prediction and ground truth, respectively.
\begin{equation}
\label{token-based-f1}
\mathrm{F1} = \frac{2 \times |\mathrm{Pred} \cap \mathrm{GT}|}{|\mathrm{Pred}| + |\mathrm{GT}|}    
\end{equation}

For LLM-as-a-Judge, we use the prompt in Appendix \ref{appendix-llm-prompt}. \textbf{Note that when using LLM-as-a-Judge, bio-entities are extracted automatically by LLMs}, so the high correlation between Entity-BLEU and LLM-as-a-Judge can illustrate the validity of our proposed bio-entity database. Results are shown in Table \ref{tab:entitybleu_multi_metric}. For both metrics, Entity-BLEU has a high correlation with them and is slightly stricter.


\subsection{Ablation Studies}

\begin{figure}
    \centering
    \includegraphics[width=.82\linewidth]{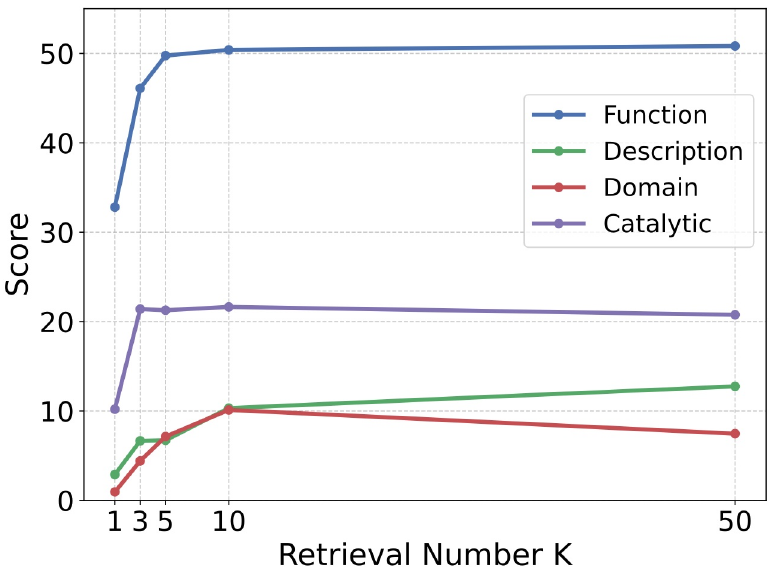}
    \vspace{-0.3cm}
    \caption{Impact of Retrieval Number $K$ in Entity-BLEU-2, randomly selected $256$ samples for each task.}
    \label{fig:ablation:k}
    \vspace{-0.4cm}
\end{figure}

\paragraph{Effect of Retrieved Sample Number.} 
We investigate the impact of varying the number of samples retrieved ($K \in \{1, 3, 5, 10, 50\}$) in the RAG pipeline. As shown in Figure ~\ref{fig:ablation:k}, increasing K can improve model performance to some extent when the number of retrieved items is small. However, a larger K may introduce low-confidence incorrect samples, thereby degrading model performance.  

\begin{table}[!htbp]
\centering
\renewcommand{\arraystretch}{0.9}
\setlength{\tabcolsep}{6pt} 
\resizebox{\linewidth}{!}{
\begin{tabular}{lccc}
\toprule
\rowcolor{gray!20}
RAG setting   & E-BLEU-2 & E-BLEU-4 & Rouge-L  \\
\midrule
GPT-4.1 w. RAPM & 56.2 & 51.4 & 34.7 \\
- w/o. Seq Index   & 50.0 (\textcolor{red}{$\downarrow$11\%}) & 44.0 (\textcolor{red}{$\downarrow$14\%}) & 31.1 (\textcolor{red}{$\downarrow$10\%}) \\
- w/o. HNSW Index  & 44.8 (\textcolor{red}{$\downarrow$20\%}) & 41.2 (\textcolor{red}{$\downarrow$20\%}) & 32.3 (\textcolor{red}{$\downarrow$7\%}) \\
- w/o. Feature Aggr.  & 51.7 (\textcolor{red}{$\downarrow$8\%}) & 47.2 (\textcolor{red}{$\downarrow$8\%}) & 31.0 (\textcolor{red}{$\downarrow$11\%}) \\
- w/o. Few-shot & 46.7 (\textcolor{red}{$\downarrow$17\%}) & 41.4 (\textcolor{red}{$\downarrow$19\%}) & 26.4 (\textcolor{red}{$\downarrow$24\%}) \\
\bottomrule
\end{tabular}
}
\caption{Impact of Retrieval Indexing Methods. We randomly selected 256 samples from the "Protein Function" task and used GPT-4.1 to generate responses.}
\label{tab:ablation:indexing}
\vspace{-0.2cm}
\end{table}
\paragraph{Effect of Database Index Methods.}


We conduct ablation studies on different database components, specifically analyzing the impact of removing: (1) the Sequence Index, (2) the HNSW Index, (3) Feature aggregation, and (4) the Few-shot component in prompts. Note: When removing the HNSW Index, Meta-Features are also eliminated.
The results (Table \ref{tab:ablation:indexing}) show that removing any index significantly affects retrieval accuracy, while the observed ROUGE-L degradation confirms the importance of Few-Shot examples for guiding LLMs to learn proper response formats.

\begin{table}[h]
\centering
\small

\setlength{\tabcolsep}{3pt}
\renewcommand{\arraystretch}{0.9}
\resizebox{\linewidth}{!}{
\begin{tabular}{lcccccc}
\toprule
\rowcolor{gray!20}
 Conf. & $<$60 & 60-70 & 70-80 & 80-90 & 90-100 & Overall \\
\midrule
Function     & 33.1 & 34.9 & 40.2 & 38.2 & \textbf{43.1} & 35.8 \\
Description  & 11.4 & 25.4 & 22.1 & \textbf{29.4} & 26.2 & 24.7 \\
Domain       & 24.0 & 23.8 & 34.6 & 27.8 & \textbf{37.0} & 26.0 \\
Catalytic    & 24.5 & 26.7 & \textbf{34.3} & 29.3 & 27.3 & 27.8 \\
\bottomrule
\end{tabular}
}

\caption{Entity-BLEU(\%) bewteen different confindence level, \textbf{Bold} for best performance}
\vspace{-4mm}
\label{tab-ablation-confidence}
\end{table}
\paragraph{Effect of Retrieved Confidence Levels.}

We analyze RAPM’s performance with different average confidence scores of the retrieved results. The confidence level is defined as the average retrieval similarity score of all recalled samples. As shown in Table \ref{tab-ablation-confidence}, while higher confidence generally leads to better performance, RAPM maintains reasonable scores even when only low-confidence (average <60\%) samples are retrieved.


\vspace{-1mm}
\section{ Conclusion and Future Works}
\vspace{-1mm}

In this work, we conduct a comprehensive analysis of existing text-based protein understanding benchmarks and methods, revealing that current benchmarks suffer from severe data leakage and that training-free retrieval-based approaches outperform fine-tuned LLM methods. 
Building on this, we introduce a novel hybrid benchmark and propose retrieval-augmented protein modelling. Our RAG method leverages both retrieval capabilities and LLMs' strengths to synthesize instruction-specific answers from retrieved evidence, achieving impressive results on OOD datasets.

Our findings highlight the effectiveness of retrieval methods for protein understanding and the need for rigorous benchmark and metric design. Future work will focus on deeper integration of retrieval and LLM methods, continuous improvements to benchmarks and metrics, and extension to other bio-entities (e.g., molecules, DNA, RNA).

\section{Acknowledgement}

This work was supported in part by Shenzhen Hetao Shenzhen-Hong Kong Science and Technology Innovation Cooperation Zone, under Grant No. HTHZQSWS-KCCYB-2023052, the Natural Science Foundation of China (No. 62202014, 62332002, 62425101), and AI4S project by PKU Shenzhen Graduate School.
\section*{Limitations}
This work primarily addresses text-based protein understanding. Extending our proposed RAG framework to other protein science tasks, such as de novo design or complex structure prediction, will require further investigation. 
The framework's effectiveness also heavily depends on the quality, coverage, and timeliness of the underlying protein knowledge database; incomplete or biased information in this resource can hinder performance, and maintaining an up-to-date database is an ongoing challenge. 
While our new benchmark and Entity-BLEU metric aim to improve evaluation rigor by mitigating data leakage and focusing on biological entities, assessing true biological understanding remains a multifaceted problem. 
Consequently, these tools, like any evaluation method, will benefit from continued validation, community adoption, and refinement. 
Furthermore, we plan to explore retrieval-augmented finetuning in future work, particularly with efficient LLMs, to further enhance domain-specific performance, an approach not investigated in this study.

\section*{Potential Risks}
A primary risk is that our framework could generate inaccurate biological insights. If unverified, these could misdirect research efforts. Over-reliance might also diminish critical human oversight. Furthermore, biases in the underlying data or LLMs could be amplified, leading to skewed predictions, especially for novel or less-studied proteins. The opaque nature of some LLMs can also make it hard to audit results or identify the root of errors. Finally, ensuring broad and equitable access to these powerful tools remains a challenge.

\bibliography{acl_latex}

\appendix

\section{Additional Results in Protein Understanding Datasets}
\label{app:more_results}

\paragraph{Performance comparison between RAPM and Prot2Text}

As shown in Table \ref{tab:prot2text-rapm}, RAPM outperforms Prot2Text$_{\text{BASE}}$ in the benchmark proposed by \cite{abdine2024prot2text}.

\begin{table}[h]
\small
\centering
\resizebox{\linewidth}{!}{
\begin{tabular}{l|c|c|c|c}
\toprule
\rowcolor{gray!25}
\textbf{Methods} & \textbf{BLEU-2} & \textbf{ROUGE-L} & \textbf{E-BLEU-2} & \textbf{METEOR} \\
\midrule
Prot2Text      & 35.1 & 48.4 & \textbackslash{} & \textbackslash{} \\
RAPM   & 47.9  & 55.4  & 38.0 & 56.6 \\
\bottomrule
\end{tabular}
}
\vspace{-0.2cm}
\caption{RAPM performance (\%) on benchmark in Prot2Text~\cite{abdine2024prot2text}.}
\label{tab:prot2text-rapm}
\end{table}

\paragraph{Performance comparison across different retrieval modalities}
The following Table~\ref{tab:retrieve_cross} shows the performances of retrieval by different protein modalities. The combined Sequence+Structure (FoldSeek+MMseq2) approach outperforms the pure Sequence (MMseqs2) retrieval for Domain and Catalytic tasks. These results suggest that structure data is a valuable complement to sequence data, indicating that a combined approach is a promising direction for future research.

\begin{table}[!h]
\centering
\small
\renewcommand{\arraystretch}{0.9}
\setlength{\tabcolsep}{3pt} 
\resizebox{\linewidth}{!}{
\begin{tabular}{lccccc}
\toprule
\rowcolor{gray!20}
Modality (Retrieval)   & Function & Description & Domain & Catalytic \\
\midrule
Sequence (MMSeqs2) & \textbf{45.2} & 16.5 &12.7 &12.6 \\
\midrule
ESM2-Embed (kNN) &39.9 &16.8 &17.5 &15.7 \\
\midrule
Structure (FoldSeek) &36.5 &14.1 &13.9 &14.5 \\
\midrule
Structure+Sequence & \multirow{2}{*}{43.7} & \multirow{2}{*}{\textbf{17.1}} &\multirow{2}{*}{\textbf{17.7}} & \multirow{2}{*}{\textbf{19.9}} \\
(FoldSeek+MMSeq2)& & & & \\
\bottomrule
\end{tabular}
}
\vspace{-0.2cm}
\caption{Performance of retrieval-based methods of different modalities.}
\label{tab:retrieve_cross}
\vspace{-0.3cm}
\end{table}


\section{LLM-as-a-Judge Prompt}
\label{appendix-llm-prompt}
For the LLM-as-a-Judge metric, we use the following prompt to evaluate the accuracy score between the prediction and the ground truth.
\begin{tcolorbox}[
    enhanced,
    colback=promptbackground,               
    colframe=black!15,                      
    coltext=prompttext,                     
    boxrule=0.5pt,                          
    arc=3mm,                                
    boxsep=5pt,                             
]
\ttfamily\small 

\textbf{System Prompt}

\vspace{1.5mm} 

Please act as a biomedical text evaluator. Follow these rules:

\begin{enumerate}[
    leftmargin=*,                                 
    label=\arabic*.\textbf{\color{prompttext}}, 
    itemsep=0.2em,topsep=1.5mm                                
]
    \item \textbf{Input:} The ``predicted\_answer'' and ``ground\_truth''.
    
    \item \textbf{Task:} Extract all bio-entities from both texts and compute a match score (0-100).
    
    \item \textbf{Scoring}:
        \begin{itemize}[
            leftmargin=1.5em,                       
            label=-,                                
            itemsep=0.2em
        ]
            \item \textbf{Perfect match} (identical entity sets): 100
            \item \textbf{Partial match}: Use the formula:
        \end{itemize}
        
        \[
        \text{Score} = \frac{2 \times \lvert E_{\text{pred}} \cap E_{\text{ref}} \rvert}{\lvert E_{\text{pred}} \rvert + \lvert E_{\text{ref}} \rvert} \times 100
        \]
        Where $E_{\text{pred}}$ and $E_{\text{ref}}$ are entities in ``predicted answe'' and ``ground truth'', respectively.

    \item \textbf{Output:} Only return an integer (0-100). No explanations.
\end{enumerate}


Evaluate: \\
predicted\_answer=\$\{predicted\_answer\}  \\
ground\_truth=\$\{ground\_truth\} \\
Output:

\end{tcolorbox}

\section{Efficiency Comparison among Different Methods}

{In this part, we systematically compare LLM-based, Retrieval-based, and RAG-based (RAPM) approaches across key ability and resource indicators, as shown in Table \ref{tab:approach_comparison}. Existing methods mostly require an extra fine-tuning process, while Retrieval-based and RAG-based methods do not. Besides, RAG-based methods effectively address the issue of chat-ability missing in Retrieval-based methods with well-designed prompts and pretrained LLMs.}

\begin{table*}[h]
\small
\centering
\resizebox{\textwidth}{!}{
\begin{tabular}{lcccccc}
\toprule
 \textbf{Methods} & \textbf{Pre-trained LLM} & \textbf{Fine-tuning} & \textbf{Retrieval} & \textbf{Chat ability} & \textbf{Run Locally} & \textbf{Training-free} \\
\midrule
\rowcolor{gray!20}
\multicolumn{7}{l}{\textbf{LLM-based}} \\
- BioGPT~\cite{biogpt}      & $\checkmark$ & $\checkmark$ & $\times$ & $\checkmark$ & $\checkmark$ & $\times$ \\
- BioT5+~\cite{pei-etal-2024-biot5}      & $\checkmark$ & $\checkmark$ & $\times$ & $\checkmark$ & $\checkmark$ & $\times$ \\
- FAPM~\cite{xiang2024fapm}       & $\checkmark$ & $\checkmark$ & $\times$ & $\checkmark$ & $\checkmark$ & $\times$ \\
- EvoLLaMA~\cite{liu2024evollama}    & $\checkmark$ & $\checkmark$ & $\times$ & $\checkmark$ & $\checkmark$ & $\times$ \\
- Galactica~\cite{taylor2022galactica}   & $\checkmark$ & $\checkmark$ & $\times$ & $\checkmark$ & $\checkmark$ & $\times$ \\
- ProLLaMA~\cite{lv2024prollama}    & $\checkmark$ & $\checkmark$ & $\times$ & $\checkmark$ & $\checkmark$ & $\times$ \\
\midrule
\rowcolor{gray!20}
\multicolumn{7}{l}{\textbf{Retrieval-based}} \\
- MMSeqs2~\cite{steinegger2017mmseqs2}         & $\times$ & $\times$ & $\checkmark$ & $\times$     & $\checkmark$ & $\checkmark$ \\
- ESM-2Embedding~\cite{esm2}  & $\checkmark$ & $\times$ & $\checkmark$ & $\times$ & $\checkmark$ & $\checkmark$ \\
\midrule
\rowcolor{gray!20}
\multicolumn{7}{l}{\textbf{RAG-based}} \\
- RAPM(GPT4.1)         & $\checkmark$ & $\times$ & $\checkmark$ & $\checkmark$ & $\times$    & $\checkmark$ \\
- RAPM(Llama3.3-70B)   & $\checkmark$ & $\times$ & $\checkmark$ & $\checkmark$ & $\checkmark$ & $\checkmark$ \\
- RAPM(DeepSeek-V3)    & $\checkmark$ & $\times$ & $\checkmark$ & $\checkmark$ & $\checkmark$ & $\checkmark$ \\
\bottomrule
\end{tabular}
}
\caption{{Comparison of resource requirements and capabilities for different approaches.}}
\label{tab:approach_comparison}
\end{table*}

\section{Additional Dataset Information and Results}
\label{app:dataset}

\paragraph{Dataset Statistics}

In this section, we introduce detailed dataset statistics information, such as data splitting and scale in Table \ref{tab:dataset-stats}, including the existing dataset and the newly proposed OOD dataset.

\begin{table}[!htbp]
    \centering
    \small
    \resizebox{\linewidth}{!}{
    \begin{tabular}{lccc}
        \toprule
        \textbf{Dataset / Task} & \textbf{Train} & \textbf{Validation} & \textbf{Test} \\
        \midrule
        \textbf{Swiss-Prot}             & 430,595  & 10,000 & 10,000  \\
        \textbf{ProteinKG25}            & 422,315  & 10,000 & 10,000  \\
        \midrule
        \textbf{Mol-Instructons (Protein)} & \multicolumn{3}{c}{} \\
        - Protein Function           & 110,689  &  --     & 3,494   \\
        - Catalytic Activity & 51,573   &  --     & 1,601   \\
        - Domain / Motif       & 43,700   &  --     & 1,400   \\
        - Functional Desc.   & 83,939   &  --     & 2,633   \\
        \midrule
        \textbf{OOD Datasets} & \multicolumn{3}{c}{} \\
        - Protein Function & 108,696 & -- & 5,487 \\
        - Domain/Motif      & 42,368 & -- & 2,732 \\
        - Catalytic Activity & 51,187 & -- & 1,987 \\
        - General Function  & 82,275 & -- & 4,297 \\
        \bottomrule
    \end{tabular}
    }
    \caption{Dataset statistics: number of samples for each task in the three corpora.}
    \label{tab:dataset-stats}
\end{table}

\paragraph{OOD dataset constructions}

The construction of OOD datasets involves three key steps:  
\begin{enumerate}[leftmargin=18pt]
 
\item Sequence Clustering: We cluster all sequences from both training and test sets using MMseqs2 with the command {\texttt{``mmseqs easy-cluster --cluster-mode 0 -c 0 -e 1e5 --single-step-clustering --min-seq-id 0 [all\_seqs]"}}, generating distinct sequence clusters.  
\item Cluster Partitioning: All clusters are randomly split into training (80\%) and test (20\%) clusters, with sequences from these clusters forming the respective training and test sets.  
\item Leakage Elimination: To prevent test-set samples from having direct training-set answers, we use \texttt{``mmseqs easy-search --max-accept 1 [query\_db] [target\_db]"} to query the most similar training-set protein for each test-set protein. If a test-set protein shares a label with its retrieved training-set counterpart, it is reallocated to the training set.  
\end{enumerate}

We repeat Step 3 twice to ensure minimal label overlap between the test set and retrieval results, yielding the final OOD dataset.

\section{Methods Details}

This section presents the methodology details, including the implementation of simple retrieval baselines and the construction of the protein knowledge database.

\subsection{Details of Simple Retrieval Methods}

To establish a straightforward baseline for the Text-based Protein Understanding task, we employ the simple retrieval approach using MMSeqs2, a widely adopted sequence alignment toolkit. Specifically, we utilize the \texttt{easy-search} mode of MMSeqs2 with the parameters \texttt{-e 1e5 --max-accept 1}. Here, \texttt{-e 1e5} sets a permissive E-value threshold to ensure the retrieval mechanism is recall-oriented, and \texttt{--max-accept 1} restricts the output to only the top candidate for each query. For every sample in the test dataset, we retrieve from the training dataset the most similar protein sequence based on alignment scores. The functional annotation (label) of the retrieved protein is then assigned as the predicted label for the query. This simple nearest-neighbor baseline is effective for assessing the upper bound of sequence-based function transfer.

\subsection{Details of Protein Knowledge Database Construction}

To construct a comprehensive protein knowledge database to support downstream tasks, we divide our methodology into data collection and efficient indexing phases.

\paragraph{Data Collection}

We integrated annotations from three prominent sources: PEER\cite{peer}, InterPro\cite{InterPro}, EC-GO\cite{gligorijevic2021structure}, and Mol-Instructions\cite{fang2023mol}. For InterPro, we selectively used only sequences annotated via Swiss-Prot curation, resulting in a high-quality subset with 573,230 sequences. In the EC-GO database, labels corresponding to the enzyme classification (EC) and gene ontology (GO) were merged into a unified text-based annotation to capture functional and process aspects simultaneously. For Mol-Instructions, only the ``meta-data'' field is retained as the annotation, disregarding the original class labels, to emphasize naturalistic, descriptive phrasing of protein functions.

\paragraph{HNSW-Index Construction}

We implement HNSW for efficient ANN search over protein sequence embeddings $\mathcal{V} = \{\mathbf{v}_i\}_{i=1}^N \subset \mathbb{R}^{1280}$ from ESM2-650M. For sequences sharing functional annotations, we compute aggregated embeddings:

\[
\mathbf{v}^{\text{agg}}_y = \frac{1}{|S_y|} \sum_{x \in S_y} \mathbf{v}_x, \quad S_y = \{x | \text{label}(x) = y\}
\]

The index construction involves three key steps:
\begin{enumerate}[leftmargin=10pt]
    \item \textbf{Layer Assignment}: Each vector $\mathbf{v}$ is assigned to layer $l$ via:
\[
l = \lfloor -\ln(\text{rand}(0,1)) \cdot m_L \rfloor, \quad m_L = 1/\ln(M)
\]

   \item \textbf{Hierarchical Insertion}: For each layer $l$ from top to bottom:
\[
\mathcal{E}_l(\mathbf{v}) = \text{argmin}_{\mathbf{u} \in \mathcal{N}_l(\mathbf{v}), |\mathcal{E}|=M} \|\mathbf{v} - \mathbf{u}\|_2
\]
where $\mathcal{N}_l(\mathbf{v})$ contains $ef\_construction$ nearest neighbors.

    \item \textbf{Small-World Guarantee}: Connections maintain:
\[
\|\mathbf{v} - \mathbf{u}\|_2 \leq r_l(\mathbf{v}), \quad \forall \mathbf{u} \in \mathcal{E}_l(\mathbf{v})
\]

The resulting structure achieves $O(\log N)$ search time with $O(N \cdot M)$ space complexity, balancing accuracy and efficiency for protein function retrieval. Key advantages include multi-layer acceleration, optimized neighborhood connectivity, and adaptive radius control.
\end{enumerate}

\paragraph{MMSeqs-Index Construction}

For indexing at the sequence level, we utilize the k-mer based inverted indexing scheme provided by MMSeqs2. Each protein sequence is decomposed into overlapping k-mers (subsequences of fixed length $k$). The index is then constructed as a mapping from each unique k-mer to the list of all sequences containing it. The search query is similarly tokenized and candidate sequences are retrieved by aggregating all records sharing at least one k-mer with the query. Formally, letting $\mathcal{K}(q)$ denote the set of $k$-mers in query $q$, the candidate set is given by 
\[
C(q) = \bigcup_{k' \in \mathcal{K}(q)} \text{Index}[k']
\]
This approach provides a highly efficient solution for large-scale substring and approximate matching, and is particularly effective for detecting local similarities.

\paragraph{Bio-Entity List Collection}
\label{abla:entity_bleu}
For the evaluation metric Entity-BLEU, we construct a domain-specific entity list.
This list is derived by extracting all distinctive biological terms from annotations in EC-GO, InterPro, and the ``metadata'' field of Mol-Instructions.
The curated entities span numerous key biological domains, meticulously categorized into areas such as \textsl{Molecular Biology and Biochemistry}(including nucleic acids like \textit{DNA}, proteins like \textit{polymerase}, and metabolites like \textit{ATP}), \textsl{Cell Biology} (covering organelles like \textit{mitochondrion}, processes like \textit{apoptosis}), \textsl{Bioenergetics and Metabolism} (e.g., \textit{glycolysis}, \textit{ATP synthase}), \textsl{Genetics and Genomics} (terms like \textit{gene}, \textit{codon}, \textit{RNA polymerase}), \textsl{Molecular Interactions and Signaling} (e.g., \textit{receptor}, \textit{MAPK pathway}), \textsl{Developmental Biology}, \textsl{Immunology}, \textsl{Plant Biology}, and \textsl{Microbiology}. Resulting in 11,341 unique terms across 10 different main categories. This ensures a broad yet granular representation of biological concepts.

To enhance specificity, we remove ambiguous general words (e.g., ``domain'') and common stopwords (e.g., ``for'', ``to''). This list underlies the Entity-BLEU metric, which rewards lexical overlap specifically on content-relevant biomedical entities, providing a fine-grained measurement of functional description quality.

\section{Experimental Details}
\subsection{Comparison Baselines}
For fair comparison, we select the following baselines:
\begin{itemize}[leftmargin=8pt]
    \item \textbf{BioT5+}~\cite{pei-etal-2024-biot5}: A T5 architecture model for biological and chemical tasks, improving on BioT5~\cite{biogpt} with IUPAC integration, multi-task tuning, and better numerical processing. 
    \item \textbf{Llama-3.2-1B}~\cite{llama3}: A recent multilingual LLM with strong generalization ability, tested in both zero-shot and fine-tuned configurations.
    \item \textbf{BioGPT}~\cite{biogpt}: A domain-specific generative Transformer language model pre-trained on large-scale biomedical literature. BioGPT achieves strong performance on six biomedical NLP tasks. Case studies demonstrate BioGPT's ability to generate fluent biomedical text descriptions.
    \item \textbf{Galactica}~\cite{taylor2022galactica}: The Galactica models are trained on a large-scale scientific corpus. The models are designed to perform scientific tasks, including but not limited to citation prediction, scientific QA, mathematical reasoning, summarization, document generation, molecular property prediction, and entity extraction. The models were developed by the Papers with Code team at Meta AI to study the use of language models for the automatic organization of science. 
    \item \textbf{ProLlama}~\cite{lv2024prollama}: ProLLaMA is a protein large language model, designed for multi-task protein language processing. It employs a two-stage training framework, incorporating Low-Rank Adaptation (LoRA) and Protein Vocabulary Pruning (PVP) to enhance efficiency. ProLLaMA achieves strong performance in protein sequence generation and property prediction tasks.
    \item \textbf{MMSeqs2 Retrieval}~\cite{steinegger2017mmseqs2}: MMseqs2 (Many-against-Many sequence searching) is a high-performance software suite designed for the rapid and sensitive retrieval of homologous protein or nucleotide sequences from large-scale databases. Its retrieval module employs a multi-stage search pipeline—comprising fast k-mer matching, ungapped alignment, and vectorized Smith-Waterman alignment—to efficiently identify relevant sequences while minimizing computational overhead. This approach enables MMseqs2 to achieve sensitivity comparable to BLAST, but with significantly enhanced speed.
    \item \textbf{ESM-2 Embedding KNN}~\cite{esm}: This method retrieves homologous proteins by performing K-Nearest Neighbors (KNN) search on fixed-length embeddings generated by the ESM-2 language model. By averaging residue-level embeddings, each protein sequence is represented as a single vector, enabling efficient similarity searches using cosine distance metrics. This embedding-based approach facilitates rapid identification of functionally similar proteins, even in cases of low sequence identity.
\end{itemize}

\subsection{Hyper-parameters}
\label{abla:hyper_params}

\paragraph{Finetune Settings.}

The LLM fine-tuning process utilizes hyperparameters shown in Table~\ref{tab:llm-hyperparams}. Training is conducted with DeepSpeed-enabled distributed GPUs, utilizing mixed-precision (bf16) and memory optimization techniques. For LLMs over 7 billion parameters, LoRA is used to significantly reduce memory requirements by freezing the majority of model weights and introducing lightweight low-rank updates. The cosine learning rate schedule with warm-up ensures stable convergence.

\begin{table}[!ht]
    \centering
    \resizebox{.9\columnwidth}{!}{
    \begin{tabular}{lc}
        \toprule
        \textbf{Hyper-parameter} & \textbf{Value} \\ 
        \midrule
        Learning rate for LoRA     & 1e-4 \\ 
        Learning rate for full parameter     & 4e-5 \\ 
        Batch size per device  & 2 \\ 
        Gradient accumulation steps & 8 \\ 
        LoRA rank & 8 \\
        LoRA $\alpha$ & 32 \\ 
        LoRA dropout & 0.05 \\ 
        Max sequence length    & 2048 \\ 
        Number of epochs       & 2 \\ 
        Optimizer              & AdamW \\ 
        LR scheduler type      & Cosine \\ 
        Warm-up ratio          & 0.1 \\ 
        Weight decay           & 1e-2 \\ 
        Mixed precision        & bf16 \\ 
        Gradient checkpointing & Enabled \\ 
        Devices                & 4 * RTX-A6000 \\
        Approximate training duration & 15 hours /task \\
        DeepSpeed config       & Zero-2 \\ 
        \bottomrule
    \end{tabular}
    }
    \caption{Hyper-parameter settings for finetuning.}
\label{tab:llm-hyperparams}
\end{table}


\subsection{RAG Inference Settings}

We standardize the inference hyperparameters across all evaluated LLMs (GPT-4.1, LLaMA3-70B, and DeepSeek-V3) to ensure fair comparison. The configuration is optimized for retrieval-augmented generation tasks:

\begin{table}[!ht]
\centering
\small
\setlength{\tabcolsep}{10mm}   
\resizebox{\linewidth}{!}{
    \begin{tabular}{lc}
    \toprule
    \textbf{Hyper-Parameter} & \textbf{Value} \\
    \midrule
    Temperature & 0.7  \\
    Top-p & 0.9  \\
    Max tokens & 2048  \\
    Frequency penalty & 0 \\
    Presence penalty & 0  \\
    \bottomrule
    \end{tabular}
}
\caption{Inference parameters for all evaluated LLMs. Identical settings are maintained across models except where architectural differences require variation.}
\label{tab:inference-params}
\end{table}

\label{abla:rag_settings}

\section{Case Study}

\begin{figure*}[!htbp]
    \centering
    \includegraphics[width=\textwidth]{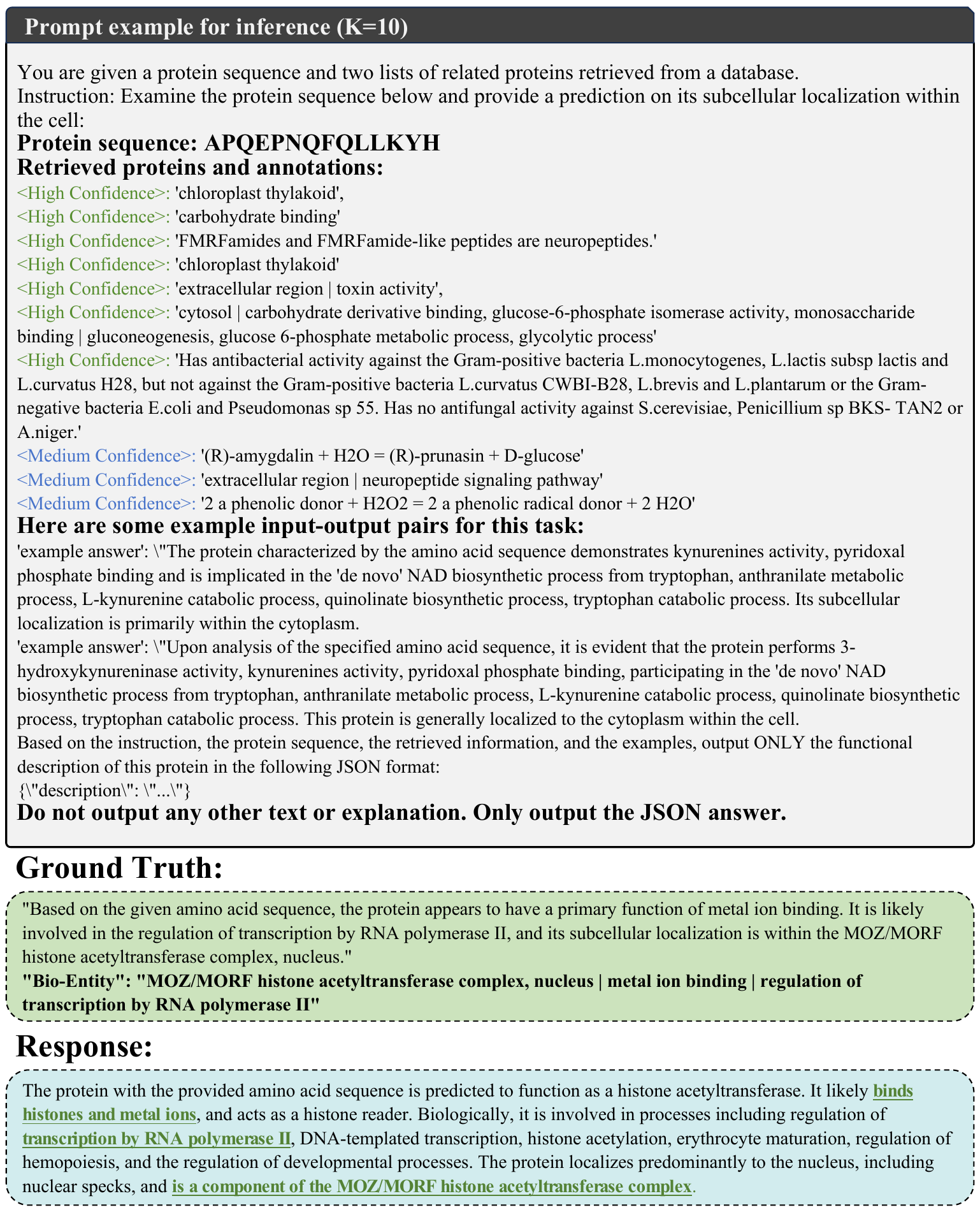}
    \caption{Example of retrieval-augmented prompt for protein knowledge injection at inference. We highlight the matching part with the ground truth in \textcolor{green!35!black}{darkgreen}.}
    \label{fig:prompt_example}
\end{figure*}

\subsection{Case Study for Data Leakage}
This part provides specific examples illustrating the data leakage observed in existing protein-text benchmarks, as discussed in Sec~\ref{sec: leakage}. Table~\ref{tab:protein_comparison} presents two representative pairs of entries, each consisting of a protein from a test dataset and a highly similar protein from its corresponding training dataset. Crucially, for both pairs, the associated functional or domain information is identical. 

For instance, the first pair shows test protein UniProt A4WLK4 and training protein UniProt A0A823T310 possessing significantly similar amino acid sequences and precisely the same detailed functional annotation (6,7-dimethyl-8-ribityllumazine synthase activity, etc.). Similarly, the second pair, UniProt Q27996 (test) and UniProt P51782 (training), exhibits high sequence homology directly correlated with an identical domain annotation ("Contains C-type lysozyme domains"). 

This direct correspondence between highly similar sequences and identical labels across the training and test sets demonstrates significant data contamination, allowing models to perform well by pattern-matching or retrieving based on superficial sequence similarity rather than developing genuine biological understanding. These cases underscore the severity of the leakage issue and motivate the need for our proposed benchmark splits.

\begin{table*}[t]
\small
\centering
\renewcommand{\arraystretch}{1.5}
\begin{tabular}{>{\raggedright\arraybackslash}p{2.6cm}|>{\raggedright\arraybackslash}p{6.5cm}|>{\raggedright\arraybackslash}p{5.2cm}}
\toprule
\textbf{Dataset} & \textbf{Protein Sequence} & \textbf{Functional/Domain Information} \\
\hline

\textbf{Test Dataset} (UniProt A4WLK4) & {\small MVRIAVVVSEFNYDVTQLMLQKALE HAKFLGAEVTYVVKVPGVYDIPTLL RDLVAKEEVDAVVTLGAVIQGATKH DEVVAHQAARKILDISVESGKPITL GIIGPGANRMQALERVEEYAKRAVE AAVKLARRKKTLREAKYAGSTVFID} & 6,7-dimethyl-8-ribityllumazine synthase activity; involved in riboflavin biosynthetic process; subcellular localization: riboflavin synthase complex. \\
\hline

\textbf{Training Dataset} (UniProt A0A832T310) & {\small MVRLAIVVAEFNYDITQLMLQKAVE HAKFLGAEITYIVKTPGVYDIPMIL KELVAKEEVDAVATLGAVIQGATKH DELVATQAARKILDIAVESGKPITL GIIGHGANRIQALERVEEYARRAVE AAVKMARRKKALREAKYNGSTVYID} & 6,7-dimethyl-8-ribityllumazine synthase activity; involved in riboflavin biosynthetic process; subcellular localization: riboflavin synthase complex. \\
\midrule

\textbf{Test Dataset} (UniProt Q27996) & {\small MKALLILGLLLLSVAVQGKTFKRCE LAKTLKNLGLAGYKGVSLANWMCLA KGESNYNTQAKNYNPGSKSTDYGIF QINSKWWCNDGKTPKAVNGCGVSCS ALLKDDITQAVACAKKIVSQQGITA WVAWKNKCRNRDLTSYVKGCGV} & Contains C-type lysozyme domains (based on computational analysis). \\
\hline

\textbf{Training Dataset} (UniProt P51782) & {\small MKVLLLLGFIFCSMAAHGKRMERCE FARRIKQLHLDGYHQISLANWVCLA QWESGFDTKATNYNPGDQSTDYGIL QINSHYWCDDGKTPHAANECKVRCS ELQEDDLVKAVNCAKKIVDQQGIRA WVAWRNKCEGKDLSKYLEGCHL} & Contains C-type lysozyme domains (based on computational analysis). \\
\bottomrule

\end{tabular}
\vspace{-0.1cm}
\caption{Comparison of Protein Sequences and Functional/Domain Annotations in Test and Training Datasets.}
\label{tab:protein_comparison}
\vspace{-0.3cm}
\end{table*}

\subsection{Case Study for RAG methods}
Fig.~\ref{fig:prompt_example} demonstrates our prompting structure, illustrating how we augment the protein sequence and query with explicit biological information retrieved from our dual-indexed database (including feature and sequence similarity results with confidence). Few-shot examples are also incorporated to guide the LLM's response format. The LLM synthesizes this retrieved evidence to generate a detailed answer about the protein. As shown, this augmented approach guides the LLM to accurately identify key biological entities and functional details compared to the ground truth, demonstrating how retrieving relevant biological context improves performance.

\section{Details on Metrics}
\label{detail_metrics}
We evaluate the model using several commonly used evaluation metrics adapted to protein description generation and understanding tasks. Here, we detail these metrics, including their calculation method, significance, and specific usage.


\textbf{BLEU:} \cite{papineni2002bleu} BLEU, or BiLingual Evaluation Understudy, is a metric often used to measure the fluency and correspondence of machine-generated sequences against reference descriptions. Employing $n$-grams, we compute the overlap:
$$
\text{BLEU} = \text{BP} \cdot \exp\left(\sum_{n=1}^N w_n \log p_n \right),
$$
where BP is a brevity penalty, $w_n$ are the weights typically equal for all $n$-grams, $\sum_{n=1}^N w_n = 1$, and $p_n$ is the precision for $n$-grams.

\textbf{ROUGE:} \cite{lin2004rouge} Recall-Oriented Understudy for Gisting Evaluation (ROUGE) measures the quality of machine-generated text by comparing its overlap with a reference set of word sequences. Specifically, it evaluates:
\begin{itemize}
    \item ROUGE-N (e.g., ROUGE-1, ROUGE-2): Measures $n$-gram overlap.
    \item ROUGE-L: Based on the longest common subsequence, it considers both recall and precision to compute an F1 score.
\end{itemize}


\section{Discussion on Licensing}
\subsection{Pretrained Models and Codes.}
\paragraph{Llama 3}\cite{llama3} The Llama 3 model is released under the Llama Community License. This license permits use, modification, and distribution, with specific conditions such as prohibitions against using the model for training other language models. For commercial use, compliance with Meta's Acceptable Use Policy is mandatory, and entities with over 700 million monthly active users must obtain a separate license from Meta.

\paragraph{GPT-4.1}\cite{gpt41} The content and models are provided under OpenAI's Terms of Use and API License Agreement. Commercial use, redistribution, or modification of GPT-4.1 models via the API requires compliance with OpenAI's policies, including attribution and adherence to usage restrictions. For full details, review OpenAI's official terms.

\paragraph{DeepSeek-V3}\cite{liu2024deepseek} DeepSeek V3 is distributed under the DeepSeek License (v1.0, Oct 23, 2023). It grants a free, global, irrevocable license for modification and distribution, with strict restrictions on military use, harm, misinformation, discrimination, and unauthorized data processing. Users must enforce these limits in derivative works. Disclaimers of warranties and liability are included, and any legal matters are subject to the jurisdiction of Chinese law, specifically in Hangzhou.

\paragraph{BioGPT}\cite{biogpt} BioGPT is released under the MIT License, permitting unrestricted use, modification, and distribution of the software and its pre-trained models, provided that the original copyright notice and license terms are included in all copies or substantial portions of the software. The software is provided "as is," without warranty of any kind, express or implied.

\paragraph{BioT5+ Model}\cite{pei-etal-2024-biot5} The BioT5+ model is available under the MIT License. This allows for free use, modification, and distribution, including for commercial purposes, as long as the original copyright notice and permission notice are retained. The software is provided "as is," with no warranties or guarantees, and the authors disclaim liability for any issues arising from its use.

\paragraph{Galactica}\cite{taylor2022galactica} The model is licensed under a non-commercial research license. This license permits use of the model and its derivatives solely for non-commercial research and evaluation purposes. Commercial use, including but not limited to using the model or its derivatives in a product or service, is strictly prohibited. Redistribution of the model weights or modifications thereof is allowed only with appropriate attribution and under the same terms. The model is provided "as is," without warranty of any kind, express or implied, including but not limited to warranties of merchantability, fitness for a particular purpose, and noninfringement.

\paragraph{ProLlama}\cite{lv2024prollama} The model is released for research and educational purposes only. Redistribution and use in source and binary forms, with or without modification, are permitted for non-commercial use provided that the original authors are properly cited. Any commercial use or use of the model or its derivatives in a commercial product or service is strictly prohibited.

\paragraph{ESM-2}\cite{esm2} ESM Metagenomic Atlas (also referred to as “ESM Metagenomic Structure Atlas” or “ESM Atlas”) data is available under a CC BY 4.0 license for academic and commercial use. Copyright (c) Meta Platforms, Inc. All Rights Reserved. 

\paragraph{MMSeqs2}\cite{steinegger2017mmseqs2} MMseqs2 is licensed under the MIT License, permitting free use, modification, and distribution of the software, provided that the original copyright notice and license terms are included in all copies or substantial portions of the software. The software is provided "as is," without warranty of any kind, express or implied.

\subsection{Datasets}
\paragraph{UniProt Database}\cite{uniprot2019uniprot} The UniProt Database is available under the Creative Commons Attribution (CC BY 4.0) License. This license permits users to share and adapt the data for any purpose, provided appropriate credit is given, a link to the license is provided, and an indication of any changes made is specified.

\paragraph{Mol-Instructions Dataset}\cite{fang2023mol} Released under the Creative Commons Attribution-NonCommercial 4.0 International License (CC BY-NC 4.0). This license permits use, sharing, and adaptation of the dataset for non-commercial purposes, with appropriate attribution and indication of changes. Commercial use requires additional permissions.
\label{sec: appendix}

\end{document}